\PassOptionsToPackage{table}{xcolor}
\documentclass{fairmeta}

\makeatletter
\providecommand{\@LN}[2]{}
\providecommand{\@LN@col}[1]{}
\makeatother

\usepackage{amsmath,amssymb}
\usepackage{tabularx}
\usepackage{array}
\usepackage{pifont}
\usepackage{url}
\usepackage{tikz}
\usetikzlibrary{arrows.meta,positioning,fit}
\definecolor{citationpurple}{HTML}{660874}
\hypersetup{colorlinks=true,citecolor=citationpurple,linkcolor=black,urlcolor=black,filecolor=black,pdftitle={FrontierCowork: Benchmarking LLM Agents at the Frontier of Real-World Professional Work},pdfauthor={Zheng Jiang, Fang Niu, Yong Yan, Fangzhou Xiong, Xinjie Lin, Yuzhao Peng, Deyao Hong, Houde Qian, Huan-ang Gao, Lifeng Sun},pdfsubject={}}

\let\cite\citep
\newcommand{\modellogo}[1]{\raisebox{-0.22em}{\includegraphics[height=1.05em]{figures/logos/#1.png}}\hspace{0.35em}}
\definecolor{overallbg}{HTML}{FBEFE6}
\definecolor{markyes}{HTML}{2E8B57}\definecolor{markno}{HTML}{C0392B}\definecolor{markpart}{HTML}{C28A2C}
\newcommand{\cmark}{\textcolor{markyes}{\ding{51}}}
\newcommand{\xmark}{\textcolor{markno}{\ding{55}}}
\newcommand{\pmark}{\textcolor{markpart}{$\circ$}}
\newlength{\singlecol}
\title{FrontierCowork: Benchmarking LLM Agents at the Frontier of Real-World Professional Work}
\author[*]{Zheng Jiang}
\author[*]{Fang Niu}
\author{Yong Yan}
\author{Fangzhou Xiong}
\author{Xinjie Lin}
\author{Yuzhao Peng}
\author{Deyao Hong}
\author{Houde Qian}
\author{Huan-ang Gao}
\author[\dagger]{Lifeng Sun}
\affiliation{Tsinghua University}
\contribution[*]{Equal contribution}
\contribution[\dagger]{Corresponding Author}

\metadata[Code]{\url{https://github.com/nf202/FrontierCowork}}
\metadata[Contact]{\href{mailto:jz24@mails.tsinghua.edu.cn}{jz24@mails.tsinghua.edu.cn}\quad\href{mailto:sunlf@tsinghua.edu.cn}{sunlf@tsinghua.edu.cn}}
\date{\today}
\abstract{
LLM agents are increasingly asked to carry out professional work rather than answer questions about it.
Real assignments often require agents to continue existing work under evolving requirements and coordinate information across heterogeneous resources.
Existing benchmarks capture these demands only partially, leaving the limitations of current agents insufficiently understood.
We characterize six cowork challenges along temporal and spatial axes: work continuation, long-horizon execution, dynamic adaptation, requirement discovery, cross-resource dependency, and multimodal grounding.
We present FrontierCowork, a benchmark of 120 tasks spanning office documents, communication, research, engineering, business operations, and media design.
Each task is constructed from expert requirements and authentic materials and requires a usable professional deliverable, such as a working spreadsheet or an editable design project.
Reproducible environments and validated scoring procedures support consistent comparisons across systems, with hidden verifiers assessing deliverables and independent model-based review reserved for qualities that cannot be checked programmatically.
Across ten frontier systems, the strongest achieves only 58.2 out of 100 and remains below 60 in four of the six domains.
}

\begin{document}
\maketitle
\pagestyle{plain}
\thispagestyle{empty}

\section{Introduction}
\label{sec:introduction}

\begin{figure*}[t]
\centering
\includegraphics[width=\textwidth]{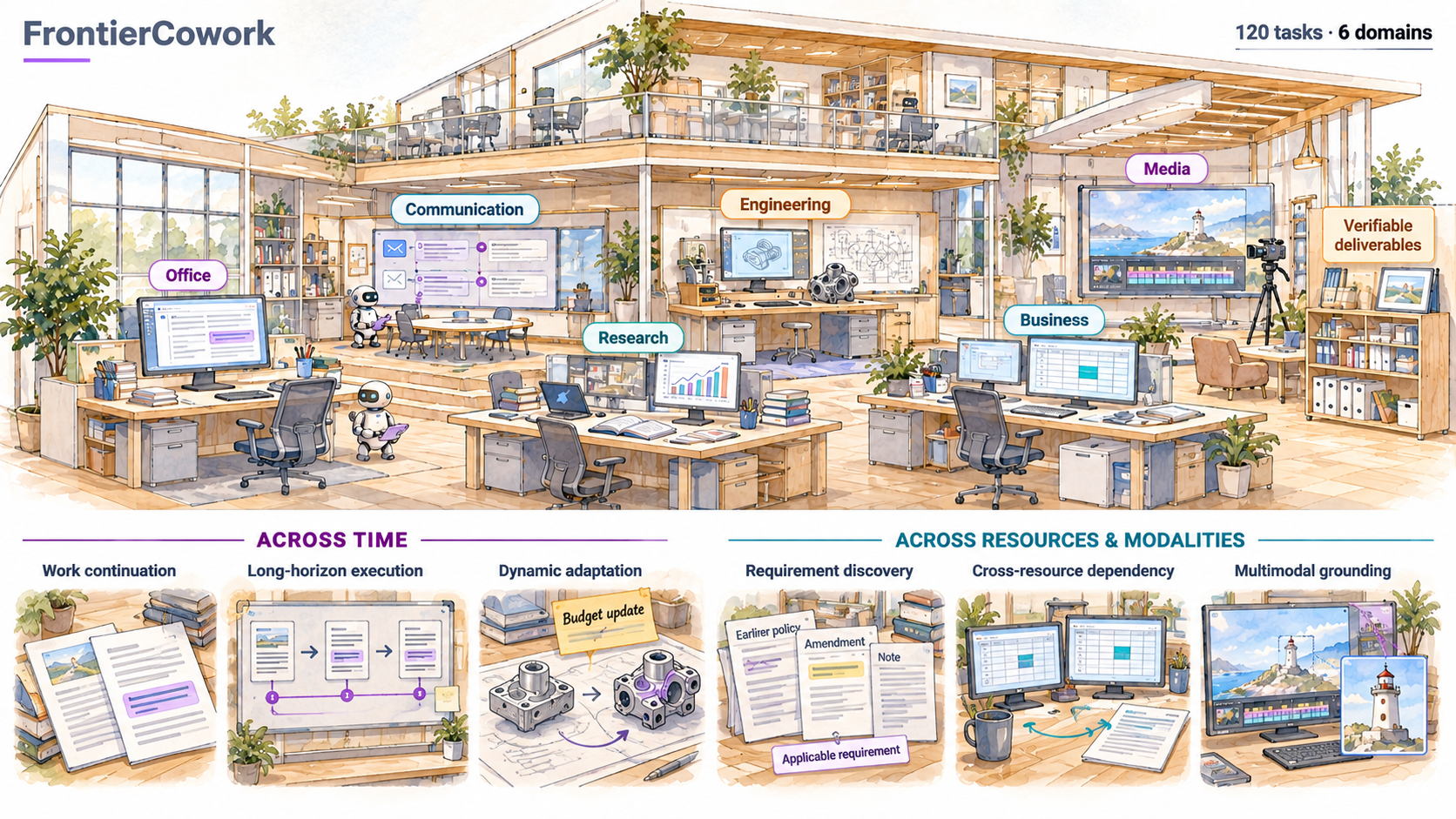}
\caption{Overview of FrontierCowork, comprising 120 tasks across six professional domains. The upper scene illustrates the work settings and deliverables in office work, communication, research, engineering, business operations, and media design. The lower panels illustrate six challenges: work continuation, long-horizon execution, and dynamic adaptation across time; and requirement discovery, cross-resource dependency, and multimodal grounding across resources and modalities.}
\label{fig:taxonomy}
\end{figure*}

With the rapid advancement of large language models, AI systems are shifting from answering questions to directly completing work.
A large share of office and knowledge work, such as reconciling financial workbooks, coordinating projects over email, auditing statistical claims, repairing engineering artifacts, and producing design deliverables, is standardized and carried out entirely through software, and it can therefore in principle be delegated to autonomous agents.
Cowork agents, which understand a goal, plan the work, operate real applications, and deliver usable results, have consequently emerged as a promising bridge between foundation models and everyday productivity, and they have attracted sustained investment from leading technology companies.
Unlike coding agents, whose progress has been driven by executable environments and verifiable benchmarks mined from real software repositories~\cite{jimenez2024swebench}, the technical route toward cowork agents has not yet converged, and the benchmarks needed to measure their capabilities remain underdeveloped.
This raises a central research question: {\rmfamily\bfseries where do current agents fall short when they are asked to carry out realistic professional work, and how can these shortcomings be measured reliably?}

Analyzing how agents fail on genuine work assignments, we identify six challenges that distinguish cowork from the tasks on which agents are usually evaluated, and existing benchmarks capture them only partially because of three limitations.
(1) \textit{Missing temporal challenges}: most benchmarks start from an empty state, end within a single episode, and state all requirements at the outset~\cite{zhou2024webarena,xie2024osworld,li2026toolathlon,wu2025mcpmark}, and even recent benchmarks of professional deliverables rarely ask agents to continue an existing work product or to revise finished work when requirements change midway~\cite{patwardhan2025gdpval,sun2026ale,vidgen2026apexagents}.
(2) \textit{Missing spatial challenges}: tasks usually come with a fully specified instruction and are confined to a single application, a set of tool servers, or a short answer~\cite{mialon2024gaia,luo2025mcpuniverse,merrill2026terminalbench}, whereas real work requires discovering the actual requirement from vague commissions and scattered materials, keeping information consistent across independent files and applications, and grounding decisions in images, renders, and frames.
(3) \textit{Unverified success criteria}: many evaluations of open-ended work rely on human preference or on model-based judges~\cite{patwardhan2025gdpval,zhao2026cle,ye2026claweval}, or on programmatic checkers whose own correctness is never checked, which leaves the reliability of the score unmeasured and allows unstated requirements to determine the outcome~\cite{zhu2025abc}.
Table~\ref{tab:comparison} summarizes how recent benchmarks address these dimensions.

\begin{table}[t]
\centering
\caption{Coverage of the six cowork challenges and of verified scoring by recent benchmarks of professional and coworker agents (\cmark: covered; \pmark: partially covered; \xmark: not covered). Verified scoring denotes hidden programmatic verifiers validated against reference and null submissions.}
\label{tab:comparison}
\small
\setlength{\tabcolsep}{5pt}
\renewcommand{\arraystretch}{1.2}
\resizebox{\textwidth}{!}{%
\begin{tabular}{lccccccc}
\toprule
 & \multicolumn{3}{c}{\textbf{Temporal}} & \multicolumn{3}{c}{\textbf{Spatial}} & \\
\cmidrule(lr){2-4}\cmidrule(lr){5-7}
\textbf{Benchmark} & \shortstack{\textbf{Work}\\\textbf{continuation}} & \shortstack{\textbf{Long-horizon}\\\textbf{execution}} & \shortstack{\textbf{Dynamic}\\\textbf{adaptation}} & \shortstack{\textbf{Requirement}\\\textbf{discovery}} & \shortstack{\textbf{Cross-resource}\\\textbf{dependency}} & \shortstack{\textbf{Multimodal}\\\textbf{grounding}} & \shortstack{\textbf{Verified}\\\textbf{scoring}} \\
\midrule
GDPval~\cite{patwardhan2025gdpval} & \xmark & \pmark & \xmark & \pmark & \pmark & \pmark & \xmark \\
APEX-Agents~\cite{vidgen2026apexagents} & \xmark & \pmark & \xmark & \pmark & \cmark & \xmark & \pmark \\
Agents' Last Exam~\cite{sun2026ale} & \xmark & \cmark & \xmark & \pmark & \pmark & \cmark & \pmark \\
OfficeEditBench~\cite{wu2026officeeditbench} & \cmark & \xmark & \pmark & \pmark & \pmark & \xmark & \pmark \\
Workspace-Bench~\cite{tang2026workspacebench} & \pmark & \pmark & \xmark & \pmark & \cmark & \pmark & \pmark \\
EvoCowork~\cite{liu2026evocowork} & \pmark & \pmark & \xmark & \xmark & \cmark & \pmark & \cmark \\
ClawMark~\cite{meng2026clawmark} & \pmark & \cmark & \cmark & \xmark & \cmark & \cmark & \pmark \\
Claw-Eval~\cite{ye2026claweval} & \xmark & \pmark & \pmark & \pmark & \pmark & \cmark & \pmark \\
CLE~\cite{zhao2026cle} & \pmark & \cmark & \cmark & \pmark & \cmark & \pmark & \pmark \\
\midrule
\rowcolor{overallbg}\textbf{FrontierCowork (ours)} & \cmark & \cmark & \cmark & \cmark & \cmark & \cmark & \cmark \\
\bottomrule
\end{tabular}%
}
\end{table}

To address these limitations, we propose FrontierCowork, a frontier evaluation that targets the six cowork challenges across office documents, communication and coordination, research analysis, engineering, business operations, and media design (Figure~\ref{fig:taxonomy}).
To bring the temporal challenges into evaluation, tasks hand the agent artifacts that others have partially completed, unfold over multiple stages or rounds in which earlier results must be carried forward, and deliver new requirements or environment changes after work has begun.
To bring the spatial challenges into evaluation, tasks require agents to infer the actual requirements from commissions and supporting materials, distribute the work over independent files and applications that must stay consistent, and require the agent to inspect visual inputs and outputs.
Each task asks for the output a colleague would hand over, such as a working spreadsheet or an editable design project, and a verifier that the agent never sees checks it against the requirements stated in the task, with an independent model reviewing only the qualities that cannot be checked programmatically.

Our contributions are threefold:

(1) \textbf{Systematic Characterization of Cowork Challenges.} We identify six challenges along a temporal and a spatial axis that separate realistic professional work from existing agent evaluations, namely work continuation, long-horizon execution, dynamic adaptation, requirement discovery, cross-resource dependency, and multimodal grounding.

(2) \textbf{Benchmark Construction and Evaluation Methodology.} We present FrontierCowork, a benchmark of 120 tasks across six domains, combining 3.9 challenges per task on average. Tasks are constructed from expert requirements and real professional materials, checked by experts before release, and executed in isolated containers with identical initial states. Deliverables are assessed by hidden domain-specific verifiers to support reproducible comparisons across agents.

(3) \textbf{Empirical Analysis of Agent Capabilities and Limitations.} We evaluate ten frontier models and find that even the strongest scores only 58.2 and stays below 60 in four of the six domains. Case studies further show how failures in managing the work can undermine otherwise capable systems.

\section{The FrontierCowork Benchmark}
\label{sec:benchmark}

\begin{figure*}[t]
\centering
\includegraphics[width=\textwidth]{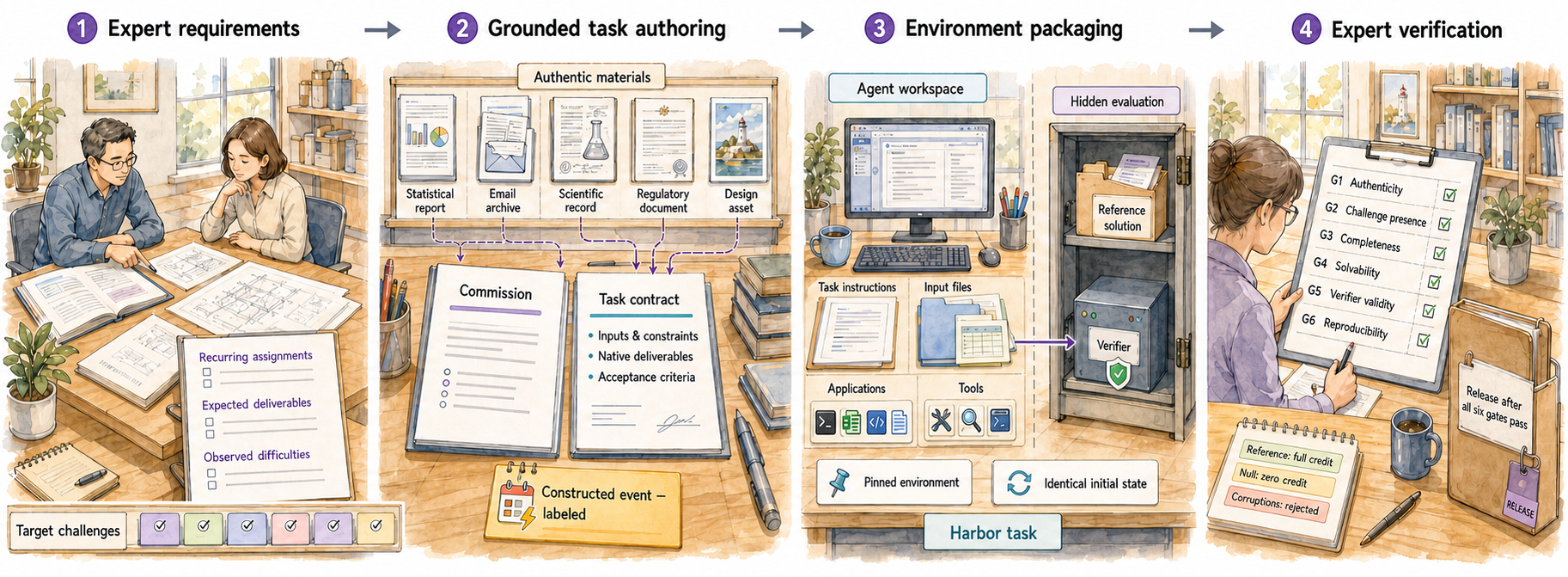}
\caption{The FrontierCowork task construction pipeline. Expert requirements and target cowork challenges guide task authoring on authentic materials. Tasks are packaged in Harbor with hidden reference solutions and verifiers, then reviewed by domain experts against six construction gates before release.}
\label{fig:pipeline}
\end{figure*}

FrontierCowork evaluates whether an agent can carry out a realistic professional assignment, starting from an initial workspace and producing a verifiable deliverable.
Its design follows four requirements: \emph{realism}, with tasks grounded in genuine work needs and authentic materials; \emph{reproducibility}, with a fixed environment and an identical initial state for each run; \emph{verifiability}, with explicit criteria for assessing deliverables; and \emph{fairness across systems}, with common task interfaces, budgets, and scoring rules across harnesses.
Within this framework, tasks target six challenges that arise in professional work.
Figure~\ref{fig:pipeline} outlines task construction, and Figure~\ref{fig:benchmark-statistics} summarizes the resulting task set.
We first define these challenges and the task formulation, then describe task construction, domain coverage, and evaluation.

\suppressfloats[t]

\subsection{Six Cowork Challenges}
\label{sec:challenges}

We organize the challenges along temporal and spatial axes; Figure~\ref{fig:benchmark-statistics}b shows their prevalence across the task set.
The temporal axis concerns how work unfolds over time.
\emph{Work continuation} requires the agent to extend or repair an existing work product while preserving the parts that must remain intact.
For example, updating one option in a government cost annex with closing actuals and a reforecast must leave the other options and source sections unchanged.
\emph{Long-horizon execution} requires the agent to maintain valid intermediate results over extended workflows in which earlier results determine later steps.
In one business task, temporary sources are withdrawn in the third of fourteen stages, making subsequent work depend on evidence saved earlier.
\emph{Dynamic adaptation} requires the agent to revise its plan and completed work in response to new requirements, corrections, or changes in the environment.
For example, an authority bulletin arriving after a baseline submission cuts the budget and withdraws approval for a heat pump, requiring revisions to the building energy simulation, controller, and procurement decision.

The spatial axis concerns how the work is distributed across information and resources.
\emph{Requirement discovery} requires the agent to determine the operative goals, rules, versions, or scope from vague commissions and potentially conflicting, stale, or unauthorized materials.
For example, two emails may jointly establish which parts of a contract can be edited.
\emph{Cross-resource dependency} requires the agent to keep information, results, and references consistent across distinct files, applications, and deliverables as upstream changes propagate.
For example, project and task records in an ERP system must remain consistent with an analysis workbook after each update; sheets within a single workbook do not count as independent resources.
\emph{Multimodal grounding} requires the agent to interpret visual inputs and inspect visual deliverables.
For example, the agent must read a reviewer's annotations on a screenshot, revise a dashboard, and render it again to check the result.

We assign a challenge to a task only when taking it away would make the task clearly easier, for example by starting from scratch instead of from inherited work, by stating every requirement up front, or by merging all resources into one file.
Across the task set, 97\% of tasks exhibit cross-resource dependency, 95\% long-horizon execution, 58\% requirement discovery, 55\% work continuation, 46\% dynamic adaptation, and 38\% multimodal grounding.
Each task exhibits 3.9 challenges on average; Appendix~\ref{sec:app-challenges} provides the definitions, easier variants, and exact counts.

\subsection{Task Formulation}
\label{sec:formulation}

A FrontierCowork task is a tuple
\begin{equation}
    \tau = (s_0, g, \mathcal{U}, \mathcal{E}, B, V),
\label{eq:task}
\end{equation}
where $s_0$ is the initial state of the workspace, $g$ is the commission stated to the agent, $\mathcal{U}$ is the public tool and action surface, $\mathcal{E}$ is the set of task artifacts and execution environments, $B$ is the time budget, and $V$ is a verifier.
The agent observes the commission and the public part of $(s_0, \mathcal{U}, \mathcal{E})$, but it never observes the verifier, the reference solution, or any hidden reference values.
Executing the task within budget $B$ yields a terminal state $s_T$ that contains the deliverable, together with an execution history $h_T$.
The verifier maps them to a score
\begin{equation}
    r = V(s_T, h_T) \in [0, 1].
\label{eq:reward}
\end{equation}
Tasks may unfold over several stages, with revised policies, additional data, or messages from colleagues released as work progresses.
Stage deliverables are preserved, and the verifier applies the domain-specific scoring rules described in Section~\ref{sec:evaluation}.
This formulation accommodates single-stage assignments, staged workflows, and multi-round coordination.

\subsection{Task Construction Pipeline}
\label{sec:construction}

\begin{figure*}[t]
\centering
\includegraphics[width=\textwidth]{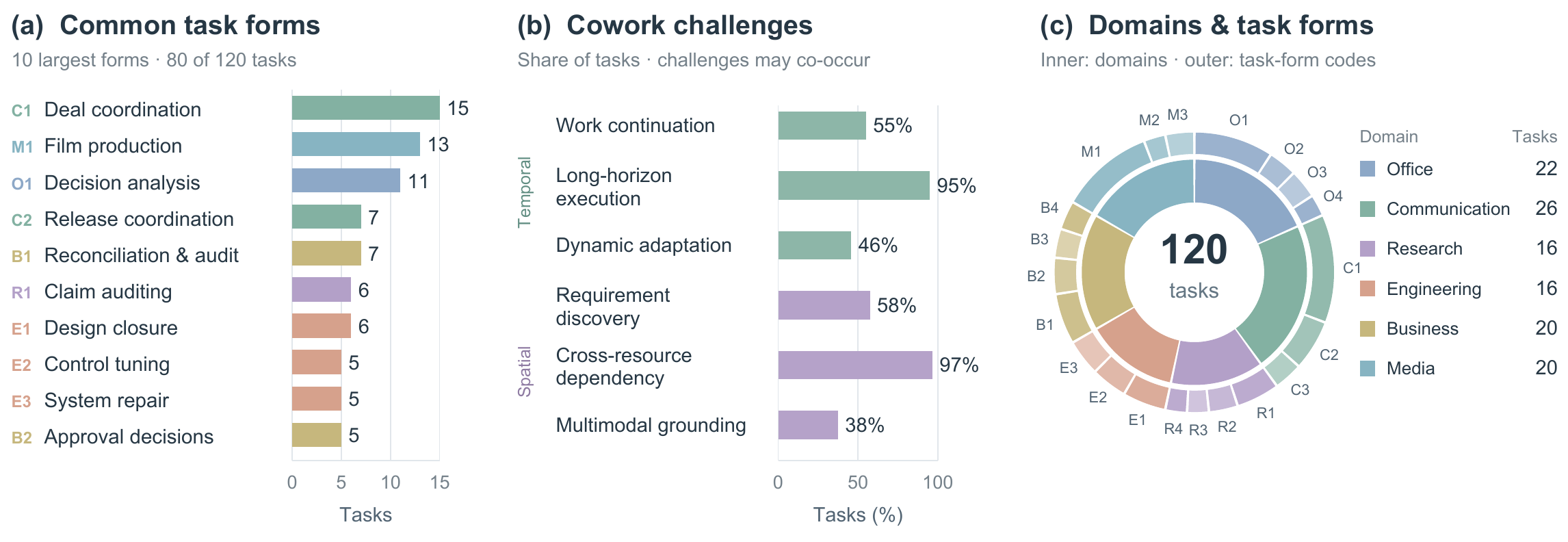}
\caption{Dataset statistics of FrontierCowork. (a) The ten largest task forms, comprising 80 of 120 tasks, colored by domain. (b) The percentage of tasks exhibiting each cowork challenge, rounded to the nearest whole percent; challenges may co-occur. (c) All 120 tasks organized by domain (inner ring) and task form (outer ring), with segment sizes proportional to task counts. Task-form codes in (a) and (c) map to the full catalog in Table~\ref{tab:catalog}; challenge annotations appear in Appendix~\ref{sec:app-challenges}.}
\label{fig:benchmark-statistics}
\end{figure*}

Tasks follow the four-stage pipeline in Figure~\ref{fig:pipeline}, from expert requirements to validated task packages.
Six construction gates translate the design requirements into explicit checks before release.

\paragraph{Collecting expert requirements.}
Domain experts describe recurring assignments, expected deliverables, and the points at which current assistants fall short.
We organize these requirements by domain and task form and assign the cowork challenges each task is intended to exercise.

\paragraph{Rewriting with authentic materials.}
Task authors turn each requirement into a concrete assignment using authentic materials, including public statistical and financial releases, open-source mailing lists and email archives, scientific measurements, procurement and regulatory data, and openly licensed design and film assets.
Each task includes a commission written from a colleague's or client's perspective and a contract specifying inputs, initial state, constraints, deliverable formats, acceptance criteria, and permitted equivalent outputs.
The commission and supporting materials together provide the basis for every assessed requirement, while identifying the applicable rules and scope may remain part of the task.
Deliverables are required in the native format of the work, such as spreadsheets with live formulas, word-processing documents, presentation files, structured records, editable design projects, and rendered media.
Controlled events, such as approvals, corrections, or policy changes, are authored on top of these materials and explicitly marked as constructed to distinguish them from source facts.

\paragraph{Environment packaging.}
Each task is packaged in the Harbor task format~\cite{harbor2026}, consisting of an agent-facing instruction, an environment specification, a reference solution, and a verifier.
The container image pins applications and dependencies, and each run restores the same initial workspace.
The verifier runs in a separate container, and the reference solution, hidden tests, and reference values are excluded from the agent-facing workspace.
Task-specific services, such as the communication domain's email workbench, run as local tool servers inside the environment.
In every domain except communication, external network access is restricted to the model service.

\paragraph{Expert verification against construction gates.}
Domain experts review each task against six construction gates; release requires passing all six.
The \emph{authenticity} gate requires recorded provenance for source materials and explicit labels for constructed events.
The \emph{challenge presence} gate requires every challenge assigned to the task to meet the criterion of Section~\ref{sec:challenges}.
The \emph{completeness} gate requires every checked requirement and permitted output equivalence to be supported by the commission, contract, or materials available to the agent at the relevant stage.
These sources need not consolidate all requirements into a single instruction: discovering which rules apply is part of the task, but hidden evaluation criteria must not introduce additional obligations.
The \emph{solvability} gate requires the reference solution to complete the task within its budget using only the inputs and tools visible to the agent.
The \emph{verifier validity} gate checks that the reference solution receives full credit and a null submission that leaves the workspace unchanged receives zero.
It also checks that permitted equivalent outputs are accepted and deliberately corrupted variants are rejected, with stage-specific verifiers validated individually.
The \emph{reproducibility} gate requires a pinned image, an identical initial state for every run, and reference materials that are hidden from the agent.

\subsection{Domains}
\label{sec:domains}

FrontierCowork comprises 120 tasks across six domains and 21 task forms (Figure~\ref{fig:benchmark-statistics}a,c), each introducing distinct professional demands.
\textbf{Office} tasks require documents to remain usable after revision, including workbooks that recompute correctly when inputs change.
\textbf{Communication} tasks require agents to maintain decisions and obligations as email archives unfold over five to nine rounds.
\textbf{Research} tasks span one or two stages, requiring evidence-based analysis and, in some cases, revisions when new evidence arrives.
\textbf{Engineering} tasks combine technical repairs with operational constraints and release decisions within a single stage, with operator events introducing changes during execution.
\textbf{Business} tasks require continuity as policies, data, and shared artifacts change across three to sixteen stages.
\textbf{Media} tasks require visual quality alongside editable native projects that support further work.
Appendix~\ref{sec:appendix} details task forms, counts, and representative assignments; Appendix~\ref{sec:app-execution} summarizes domain-specific workflows and scoring.

\subsection{Evaluation Methodology}
\label{sec:evaluation}

\paragraph{Execution.}
Each run starts from a clean container built from the pinned task image.
The agent accesses the workspace through its harness, using shell and file operations and any task-specific tools.
Task time budgets range from 20 to 180 minutes according to expected effort; business tasks are instead bounded by their stage structure, without an overall time limit.
A run receives zero if it exhausts its budget without a deliverable or produces an output the verifier cannot process.
Infrastructure failures, such as a container failing to start on a faulty node, are rerun until a valid run is obtained and do not enter the score.

\paragraph{Scoring.}
The verifier type follows the nature of each deliverable.
For artifacts with checkable content, such as spreadsheets, data tables, engineering models, and code, programmatic verifiers recompute results in the native application, compare values against reference values within stated tolerances, and check the structural and behavioral requirements of the contract.
For coordination tasks, submitted records are matched row by row against reference records supported by the available sources.
This scoring credits partial progress and rewards leaving unknown fields blank over guessing their values.
For design qualities that cannot be assessed programmatically, an independent reviewer model inspects exported media and native files using a task-specific rubric.
Each rubric has five dimensions scored from 0 to 3, including factual fidelity to the commission and sources and typically covering continuation of the native project, visual quality, and communication.
A missing deliverable receives zero.
Communication and business tasks average scores across rounds or stages.
Multi-stage research tasks use the final-stage score, since the final deliverable must incorporate the preceding stages (Appendix~\ref{sec:app-scoring}).

\paragraph{Aggregation.}
Each task receives a score $S_t \in [0, 1]$.
With $\mathcal{T}_d$ denoting the set of tasks in domain $d$, the domain score is
\begin{equation}
    \bar{S}_d = \frac{1}{|\mathcal{T}_d|} \sum_{t \in \mathcal{T}_d} S_t,
\label{eq:domain}
\end{equation}
and the overall score is the equal-weight mean of the six domain scores, so that each domain contributes equally regardless of its number of tasks.

Together, these choices address the gaps identified in Section~\ref{sec:introduction}: inherited artifacts and evolving workflows exercise temporal challenges, distributed information and visual materials exercise spatial challenges, and validated verifiers ground scores in explicit acceptance criteria.
Common execution conditions support comparisons across systems, with rubric-based review reserved for qualities that cannot be checked programmatically.

\section{Experiments}
\label{sec:experiments}

\subsection{Experimental Setup}
\label{sec:setup}

\paragraph{Evaluated systems.}
We evaluate ten frontier models, each as a complete model and harness system.
Table~\ref{tab:main} lists the systems and their harnesses; Astra runs in Codex and the other nine models run in Claude Code.
We use shortened model names in the analysis below.
Every system runs at the highest reasoning effort it supports in all six domains.
Execution and scoring details are provided in Appendix~\ref{sec:app-execution}.

\paragraph{Protocol.}
All systems are evaluated on the same 120 tasks under the methodology of Section~\ref{sec:evaluation}, with one run per task, on the Harbor execution platform~\cite{harbor2026}.
Runs that failed because of the infrastructure were rerun, so every system has a valid score on every task.
Media deliverables are assessed by GPT-6.1 Sol at high reasoning effort as the independent reviewer, with a fresh context for each deliverable and without access to the identity of the evaluated system.
We report scores on a 0 to 100 scale.

\subsection{Overall and Domain-Level Performance}
\label{sec:results}

\begin{table*}[htbp]
\caption{Main results on FrontierCowork (0 to 100). All systems run at the highest reasoning effort they support, Overall is the equal-weight mean of the six domain scores, and rows are sorted by Overall. Bold values denote the highest value in each column.}
\label{tab:main}
\centering
\small
\setlength{\tabcolsep}{5pt}
\renewcommand{\arraystretch}{1.15}
\resizebox{\textwidth}{!}{%
\begin{tabular}{lc|>{\columncolor{overallbg}}c|cccccc}
\toprule
Model & Harness & Overall & Office & Comm. & Research & Eng. & Business & Media \\
\midrule
\modellogo{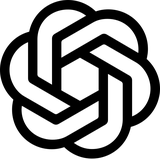}GPT-6.2 Astra & Codex & \textbf{58.2} & \textbf{59.1} & \textbf{44.4} & \textbf{73.3} & \textbf{62.5} & \textbf{57.0} & \textbf{53.0} \\
\modellogo{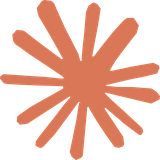}Claude Opus 5.5 & Claude Code & 49.5 & 54.5 & 33.0 & 67.0 & \textbf{62.5} & 49.4 & 30.3 \\
\modellogo{claude-color}Claude Fable 5.1 & Claude Code & 44.9 & 36.4 & 28.8 & 62.0 & 50.0 & 46.9 & 45.3 \\
\modellogo{claude-color}Claude Sonnet 5.5 & Claude Code & 41.8 & 36.4 & 31.6 & 61.8 & \textbf{62.5} & 33.9 & 24.3 \\
\modellogo{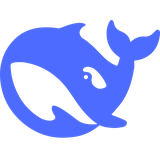}DeepSeek V4.1 Flash & Claude Code & 39.7 & 31.8 & 24.0 & 56.2 & 56.2 & 30.4 & 39.3 \\
\modellogo{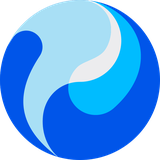}Tencent Hy4 & Claude Code & 30.5 & 22.7 & 10.0 & 49.0 & 50.0 & 30.9 & 20.0 \\
\modellogo{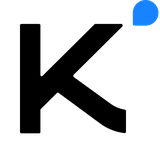}Kimi K3.1 Pro & Claude Code & 30.2 & 18.2 & 27.3 & 55.2 & 37.5 & 25.1 & 17.7 \\
\modellogo{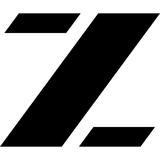}GLM-5.3 Flash & Claude Code & 27.6 & 18.2 & 21.4 & 47.4 & 25.0 & 30.9 & 22.7 \\
\modellogo{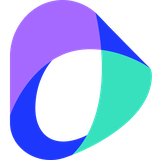}Doubao Seed Evolving & Claude Code & 25.4 & 18.2 & 13.8 & 19.9 & 50.0 & 30.9 & 19.3 \\
\modellogo{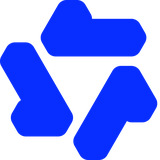}Qwen3.8 Max & Claude Code & 18.1 & 9.1 & 10.6 & 25.0 & 25.0 & 23.7 & 15.3 \\
\bottomrule
\end{tabular}%
}
\end{table*}

Table~\ref{tab:main} reports the domain and overall scores of all systems.

\paragraph{FrontierCowork leaves substantial headroom.}
The strongest system, Astra, reaches an overall score of 58.2, and no system scores above 74 in any domain.
The average across systems ranges from 24.5 in communication to 51.7 in research, indicating that substantial room for improvement remains across professional domains.

\paragraph{System rankings vary across domains.}
Astra attains the highest score in every domain and shares the lead in engineering with Opus and Sonnet at 62.5, but the order of the remaining systems changes from domain to domain.
Opus ranks second in four domains but only fourth in media design, where Fable and DeepSeek outperform it.
These variations suggest that overall performance does not fully characterize a system's capabilities in individual professional domains.

\paragraph{Domains differ in score levels and system separation.}
Communication has the lowest mean score, at 24.5, with the best system reaching 44.4.
Research and office work separate systems most strongly, with spreads of 53.5 and 50.0 points between the best and weakest systems.
These comparisons should be interpreted in light of the different scoring rules across domains.
Detailed analyses of communication task forms and media completion are provided in Appendix~\ref{sec:app-domain-results}.

\FloatBarrier
\subsection{Performance across Cowork Challenges}
\label{sec:analysis}

\begin{figure*}[htbp]
\centering
\includegraphics[width=\textwidth]{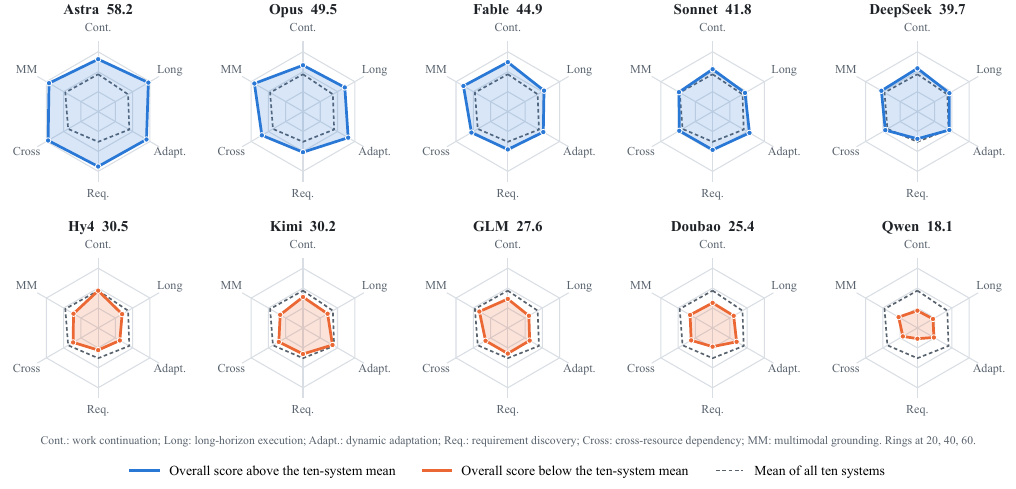}
\caption{Challenge profile of each system: the mean score (0 to 100) on the tasks that exhibit each challenge, with the overall score next to the system name. Blue marks systems whose overall score is above the mean of all ten systems and orange those below it; the dashed hexagon is the mean of all ten systems, and the full values are given in Table~\ref{tab:challenge-profile} in Appendix~\ref{sec:app-profile}.}
\label{fig:challenge-radar}
\end{figure*}

Figure~\ref{fig:challenge-radar} shows the mean score of every system on the tasks that exhibit each challenge; complete numerical results are provided in Appendix~\ref{sec:app-profile}.

Long-horizon execution and cross-resource dependency occur in almost every task, so their profiles closely follow overall performance.
The remaining challenges reveal differences between systems: Opus nearly matches Astra on multimodal grounding, at 56.3 against 57.0, and on dynamic adaptation, at 52.4 against 56.0.
On work continuation, the four leading systems lie within 10 points of one another, while DeepSeek, at 43.4, exceeds Sonnet.
Requirement discovery also distinguishes systems, with Astra scoring 55.1 compared with 38.0 to 40.8 for the other three leading systems.
Because the subsets overlap and differ in their domain composition, these profiles do not isolate the effect of a single challenge.

\FloatBarrier
\subsection{Detailed Case Studies}
\label{sec:failure-analysis}

We examine two representative tasks to show how failures emerge in delivered work.
Together they cover all six cowork challenges, although each task combines several challenges rather than isolating one.
The figures reproduce the original evidence, and the analysis traces failures to delivered files, verifier outputs, and agent trajectories.
Appendix~\ref{sec:app-cases} provides two further cases involving engineering revisions and a visual data audit.

\subsubsection{Bridgeline Coordination}
\label{sec:case-revision}
\noindent\textit{Challenges: long-horizon execution, dynamic adaptation, requirement discovery, and cross-resource dependency.}

\begin{figure*}[!t]
\centering
\includegraphics[width=\textwidth]{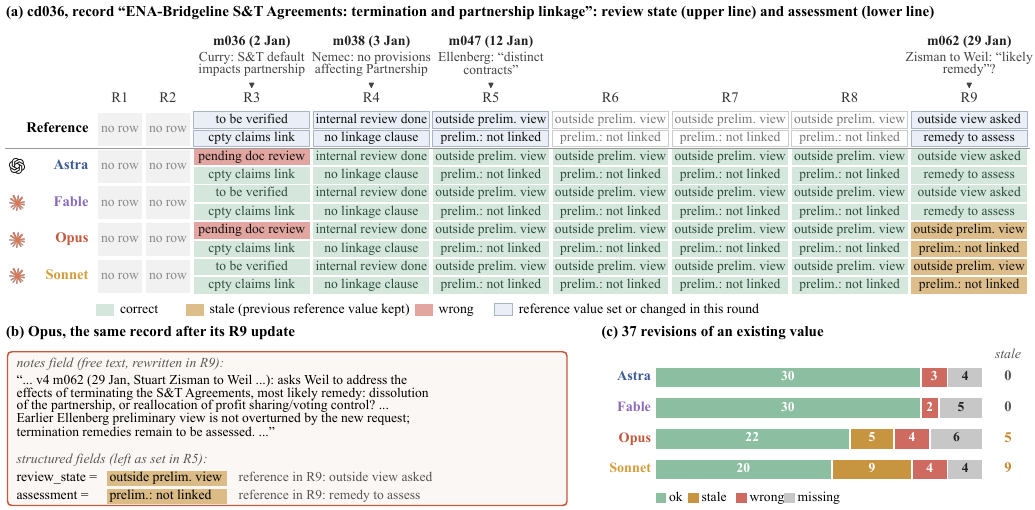}
\caption{Bridgeline Coordination (cd036). Revising finished work when late evidence reopens it.
(a) One record of cd036 over nine rounds, with the reference values on top and the values each system held below, colored by the verifier's per field result; the email that sets or changes the reference is marked above each column.
Glosses: \textit{to be verified}, \textit{pending doc review}, \textit{internal review done}, \textit{outside prelim. view} (outside counsel's preliminary assessment), \textit{outside view asked} (outside assessment requested); \textit{cpty claims link} (counterparty claims a partnership link), \textit{no linkage clause} (reviewed documents show no linkage clause), \textit{prelim.: not linked} (preliminary view that the contracts should not be linked), \textit{remedy to assess} (termination remedy still to be assessed).
(b) The R9 update of Opus changed only the free text note, which states the new open question, while both structured fields kept the R5 values.
(c) Outcome of each system on the 37 revisions of an existing reference value in cd036; stale means the submitted value equals the previous reference value.}
\label{fig:case-revision}
\end{figure*}

Communication task cd036 asks the agent to maintain five tables over nine rounds of Enron bankruptcy email and to \textit{preserve historical versions and distinguish proposals, approvals, and execution}.
The record in Figure~\ref{fig:case-revision}a tracks whether terminating ENA's storage and transport agreements would affect the Bridgeline partnership.
Its reference value settles in R5, and in R9 email m062 reopens it by asking outside counsel \textit{what is the likely remedy for such a termination?}
Because the catalog defines review state as research progress that \textit{cannot substitute for a legal conclusion}, the reference moves to an open outside assessment.

All four frontier systems held the record correctly from R4 to R8.
In R9 Astra and Fable rewrote both fields, while Opus wrote the open question into its notes and left both structured fields at its R5 values (Figure~\ref{fig:case-revision}b), and Sonnet kept them as well.
Across the 37 revisions of an existing reference value, Opus left 5 stale and Sonnet 9, against none for Astra and Fable (Figure~\ref{fig:case-revision}c).

Opus and Sonnet registered the new evidence in prose but guarded the settled field as a conclusion rather than updating it as a state, so the record a successor would filter on no longer shows the open question their own notes record.
\FloatBarrier

\subsubsection{Reading Edition for Street Project Teams}
\label{sec:case-layout}
\noindent\textit{Challenges: work continuation, long-horizon execution, cross-resource dependency, and multimodal grounding.}

\begin{figure*}[htbp]
\centering
\includegraphics[width=\textwidth]{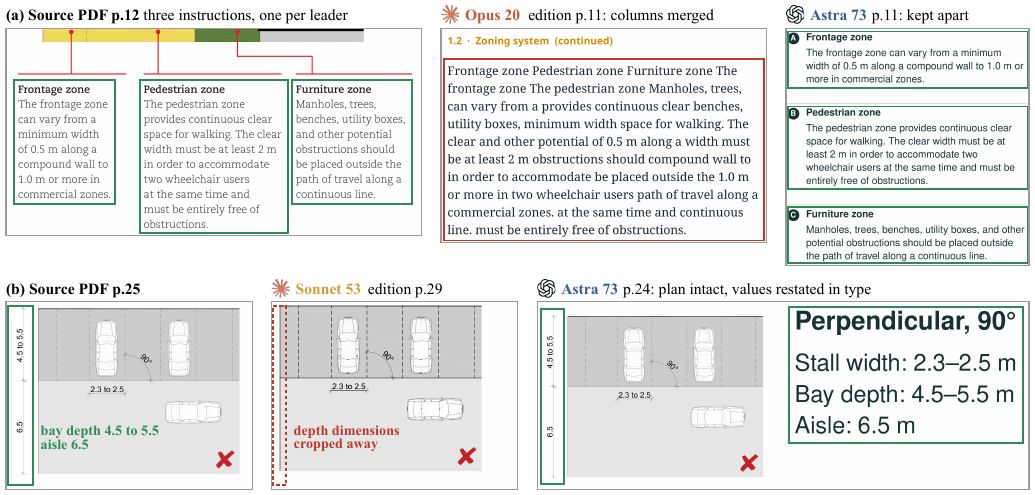}
\caption{Reading Edition for Street Project Teams (streets\_team\_\allowbreak reading\_edition). Crops from the source guide (ITDP and UN-Habitat, 2018, CC BY 4.0) and from the delivered editions, with final scores in the panel headers.
(a) Source PDF page 12 sets the frontage, pedestrian and furniture zone instructions in three columns, each tied to a leader on the cross section (green boxes).
Opus joined lines on the same baseline across the columns into one paragraph on its edition page 11 (red box), while Astra kept three separate notes keyed A to C to its drawing.
(b) The perpendicular parking plan on source PDF page 25 carries its bay depth (4.5 to 5.5) and aisle (6.5) in a dimension strip at the left.
Sonnet's crop of the plan starts to the right of that strip, so the values are gone; the angled and parallel plans lost theirs the same way.
Astra kept the strip and repeated the dimensions in live type beside the plan.}
\label{fig:case-layout}
\end{figure*}

Media task streets\_team\_reading\_edition asks for a complete reading edition of the 56 page ITDP and UN-Habitat guide \textit{Streets for Walking \& Cycling}, delivered as a PDF and an editable project, and requires that \textit{the technical content, examples and their relationships must remain intact}.
In this source, much of that meaning is carried by position: columns tied to drawing leaders, dimension strips beside plans, and photographs paired with tick and cross captions.
Every system discussed here delivered a PDF that opens and a project that passed the native edit, save and reopen test, so their factual scores turn on whether the rebuilt pages still say what the source pages say.

Opus (20) extracted text with a script that sorts lines by vertical position, which joined three zoning columns into one paragraph (Figure~\ref{fig:case-layout}a).
Sonnet (53) cropped the parking plans just right of their dimension strips, so the bay depth and aisle values disappeared (Figure~\ref{fig:case-layout}b).
Kimi (20) paired photographs with the opposite captions on six pages, and DeepSeek (20) dropped the last line of two rules.
Astra (73) keyed the zoning notes to the drawing and kept the plans with their dimensions.

None of these defects stopped a build, and Kimi's own audit of its technical paragraphs reported \textit{0 missing}.
The rubric's factual dimension scored 0 for eight of the ten systems, and the agents that failed treated extraction as finished without comparing each rendered page against its source page.
\FloatBarrier

\FloatBarrier
\subsection{Performance and Resource Efficiency}
\label{sec:resources}

\subsubsection{Performance and Token Usage}
\begin{figure*}[!ht]
\centering
\includegraphics[width=\textwidth]{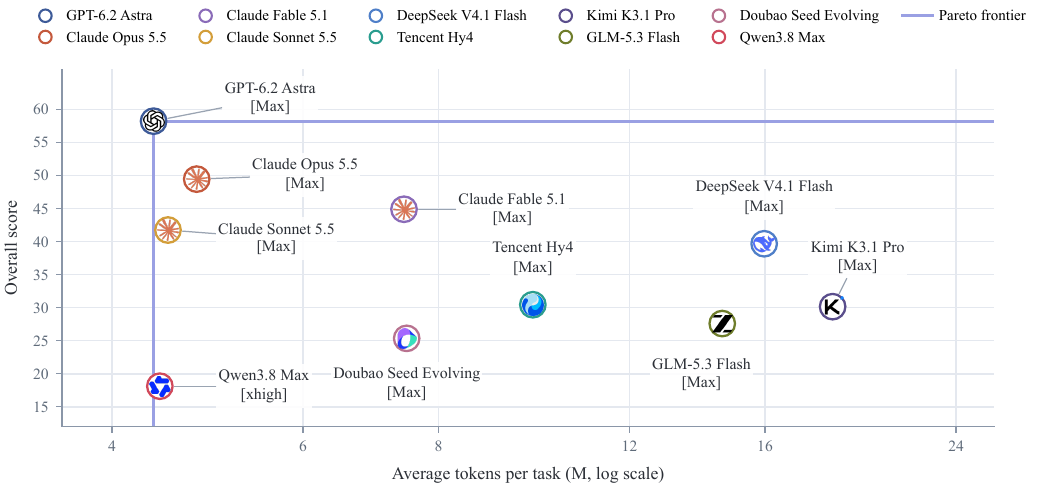}
\caption{\textbf{Overall score versus token consumption.} Overall scores match Table~\ref{tab:main}. Mean per-task tokens include uncached input, cached input, and output on all 120 tasks, normalized across providers so that cached input is counted once. The token axis is logarithmic, and the line connects non-dominated systems.}
\label{fig:resources}
\end{figure*}

Figure~\ref{fig:resources} relates overall score to mean token consumption per task.
Astra, Sonnet, Qwen, and Opus consume between 4.4 and 4.8 million tokens per task, whereas DeepSeek, GLM, and Kimi consume 14.6 to 18.5 million.
Astra lies alone on the token frontier, since no evaluated system achieves a higher score with fewer tokens.
Under these configurations, greater token consumption does not consistently correspond to higher scores; this cross-system comparison does not establish the effect of increasing the inference budget of a fixed system.

\subsubsection{Performance and Inference Cost}

\begin{figure*}[!ht]
\centering
\includegraphics[width=\textwidth]{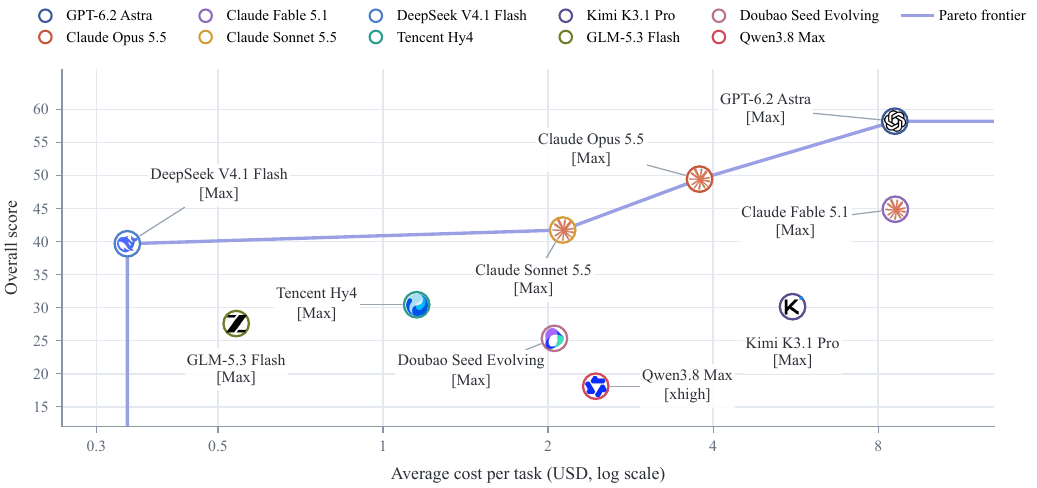}
\caption{Overall score versus mean estimated inference cost per task. Scores match Table~\ref{tab:main}; costs use recorded token usage and official list prices. The logarithmic cost axis and the line connecting non-dominated systems show the observed performance--cost frontier. Token usage, prices, and accounting assumptions are detailed in Appendix~\ref{sec:app-cost}.}
\label{fig:resources-cost}
\end{figure*}

Figure~\ref{fig:resources-cost} shows that the strongest system is not the least expensive option.
The observed cost frontier runs from DeepSeek through Sonnet and Opus to Astra.
DeepSeek achieves 39.7 at an estimated 0.34 dollars per task, Sonnet 41.8 at 2.13 dollars, Opus 49.5 at 3.78 dollars, and Astra 58.2 at 8.58 dollars.
The comparison therefore exposes a substantial trade-off between task performance and inference cost under the evaluated configurations.
These are list-price estimates rather than measured charges; cache-write costs are excluded where provider reporting is incomplete.
Fable costs about as much as Astra for a score 13.3 points lower, while Kimi costs more than Opus for a score 19.3 points lower.
The cost frontier therefore differs from the token frontier because provider prices and cache discounts vary substantially.
Appendix~\ref{sec:app-cost} gives the complete token accounting and cost estimates.

\section{Related Work}
\label{sec:related_work}

\subsection{Agent Benchmarks in Interactive Environments}
Recent advances in LLM agents have improved learning from interaction~\cite{yang2026browserskill,li2026embodiedevo,jiang2026frontieropd} and reasoning grounded in visual evidence~\cite{jiang2026medvr,jiang2026vista,jiang2026ttsp}, motivating broader evaluation of their capabilities in interactive environments.
Agent evaluation has moved from static question answering toward interactive environments with outcome-based verification, covering websites~\cite{zhou2024webarena}, operating systems~\cite{xie2024osworld,yuan2026osworld2,jia2025osworldmcp}, applications~\cite{trivedi2024appworld}, software repositories~\cite{jimenez2024swebench,yang2024sweagent}, and terminals~\cite{merrill2026terminalbench}.
Tool-use benchmarks built on the Model Context Protocol evaluate agents across many applications and tool servers from realistic initial states~\cite{li2026toolathlon,wu2025mcpmark,luo2025mcpuniverse}, while dynamic environments such as Gaia2 and the $\tau$-bench family let the world or a simulated user change the shared state during an episode~\cite{froger2025are,yao2024taubench,barres2025tau2bench}.
These benchmarks established executable and verifiable evaluation, but their tasks typically start from a clean state with a fully specified goal and end in a short answer or an environment state rather than a work product that colleagues continue to use.

\subsection{Benchmarks for Professional Work}
A growing line of work evaluates agents on economically valuable professional tasks, including complex workflow orchestration and 3D content creation~\cite{jiang2026workflowarena,lin2026frontier3dbench}.
GDPval and the Remote Labor Index collect deliverables from real occupations and freelance projects and grade them by expert comparison~\cite{patwardhan2025gdpval,mazeika2025rli}, APEX-Agents and Agents' Last Exam build long-horizon expert tasks with native artifacts~\cite{vidgen2026apexagents,sun2026ale}, and SWE-Lancer prices real software contracts~\cite{miserendino2025swelancer}.
Office-centric benchmarks study spreadsheets, documents, and workspaces~\cite{zhu2026spreadsheetbench2,wu2026officeeditbench,tang2026workspacebench,zhou2026omegauseofficeval,wang2025odysseybench}, and EvoCowork evaluates harness self-improvement on office deliverables~\cite{liu2026evocowork}.
Coworker benchmarks in the Claw family place agents in stateful multi-service workplaces~\cite{meng2026clawmark,ding2026wildclawbench,ye2026claweval,zhong2026enterpriseclawbench}, CLE measures how agents improve across sequences of professional assignments~\cite{zhao2026cle}, and WorkflowArena evaluates how agents organize multi-agent workflows on long-horizon tasks~\cite{jiang2026workflowarena}.
FrontierCowork complements these efforts by targeting six explicitly defined cowork challenges and by covering office, communication, research, engineering, business, and media work within a single task format graded by hidden verifiers.

\subsection{Verifiable and Reproducible Evaluation}
The reliability of agentic benchmarks depends on whether task setups and graders measure what they claim, and recent work has proposed a checklist for building rigorous agentic benchmarks~\cite{zhu2025abc}.
Containerized task formats such as Harbor and Terminal-Bench pair each task with an environment, a reference solution, and tests~\cite{harbor2026,merrill2026terminalbench}, and recent benchmarks validate their checkers against reference and failing submissions~\cite{jiang2026workflowarena,liu2026evocowork}, whereas evaluations of open-ended work often rely on human preference or model-based judges~\cite{patwardhan2025gdpval,zhao2026cle}.
Related work on physical consistency also emphasizes grounding evaluation in measurable evidence and validating the evaluator under controlled conditions~\cite{gao2026worldmodelslastexam}.
FrontierCowork builds on the Harbor format, pairs every task with a contract that states its checked requirements, validates every verifier against reference, null, equivalent, and corrupted submissions, and restricts model-based review to the design qualities that cannot be checked programmatically.

\section{Conclusion}
\label{sec:conclusion}

We identified six challenges that distinguish realistic professional work from existing agent evaluations, organized along a temporal axis of work continuation, long-horizon execution, and dynamic adaptation and a spatial axis of requirement discovery, cross-resource dependency, and multimodal grounding.
Targeting these challenges, we introduced FrontierCowork, a benchmark of 120 tasks across six domains in which agents work from real materials toward the output a colleague would hand over, graded by verifiers they never see.
Returning to the question posed in the introduction, current agents carry out a meaningful part of realistic professional work but remain far from completing it reliably: the strongest system reaches an overall score of 58.2.
We hope that FrontierCowork and its challenge taxonomy will guide the development of agents that can be trusted with genuine work.

\bibliographystyle{assets/plainnat}
\bibliography{references}

\clearpage
\beginappendix
\section{Task Catalog and Construction Details}
\label{sec:appendix}

\subsection{Task Forms and Representative Assignments}

Table~\ref{tab:catalog} lists the task forms within each domain, which describe what the agent must do and deliver, together with the number of tasks of each form.
Task-form codes identify the outer-ring segments in Figure~\ref{fig:benchmark-statistics}; individual tasks retain their public identifiers in the released package.

\begin{table*}[htbp]
\centering
\caption{Task forms in FrontierCowork and the number of tasks of each form.}
\label{tab:catalog}
\small
\setlength{\tabcolsep}{4pt}
\begin{tabular}{>{\raggedright\arraybackslash}p{2.3cm}>{\raggedright\arraybackslash}p{3.6cm}>{\raggedright\arraybackslash}p{8.2cm}r}
\toprule
\textbf{Domain} & \textbf{Task form} & \textbf{What the agent must deliver} & \textbf{Tasks} \\
\midrule
\multirow{4}{*}{Office} & \textsf{\scriptsize O1}\enspace Decision analysis & Workbooks and memos that support staged decisions & 11 \\
 & \textsf{\scriptsize O2}\enspace Template completion & Official templates completed in place & 4 \\
 & \textsf{\scriptsize O3}\enspace Document revision & Existing workbooks and documents repaired or revised & 4 \\
 & \textsf{\scriptsize O4}\enspace Report construction & Workbooks, documents, and slides built from source data & 3 \\
\midrule
\multirow{3}{*}{Communication} & \textsf{\scriptsize C1}\enspace Deal coordination & Records and briefs that track business transactions over email & 15 \\
 & \textsf{\scriptsize C2}\enspace Release coordination & Records and briefs that track open-source releases & 7 \\
 & \textsf{\scriptsize C3}\enspace Commitment tracking & Records and briefs that track commitments across projects & 4 \\
\midrule
\multirow{4}{*}{Research} & \textsf{\scriptsize R1}\enspace Claim auditing & Statistical claims checked against source data & 6 \\
 & \textsf{\scriptsize R2}\enspace Rule-based analysis & Indicators and assessments computed under explicit rules & 4 \\
 & \textsf{\scriptsize R3}\enspace Record linkage & Deduplicated record ledgers & 3 \\
 & \textsf{\scriptsize R4}\enspace Pipeline migration & Analysis pipelines migrated with reproduced results & 3 \\
\midrule
\multirow{3}{*}{Engineering} & \textsf{\scriptsize E1}\enspace Design closure & Engineering designs closed against constraints and change notices & 6 \\
 & \textsf{\scriptsize E2}\enspace Control tuning & Identified models and tuned controllers & 5 \\
 & \textsf{\scriptsize E3}\enspace System repair & Repaired engineering files, code, and services & 5 \\
\midrule
\multirow{4}{*}{Business} & \textsf{\scriptsize B1}\enspace Reconciliation and audit & Reconciled and audited workbooks and ERP records & 7 \\
 & \textsf{\scriptsize B2}\enspace Approval decisions & Approval and scope decisions derived from policies & 5 \\
 & \textsf{\scriptsize B3}\enspace Dashboard revision & Dashboards revised after review & 4 \\
 & \textsf{\scriptsize B4}\enspace Capacity planning & Capacity plans and scenarios in ERP systems and workbooks & 4 \\
\midrule
\multirow{3}{*}{Media} & \textsf{\scriptsize M1}\enspace Film production & Films produced or adapted from native projects & 13 \\
 & \textsf{\scriptsize M2}\enspace Publication layout & Publications relaid out for print and screen & 3 \\
 & \textsf{\scriptsize M3}\enspace Design revision & Spatial, product, photographic, and presentation designs revised & 4 \\
\bottomrule
\end{tabular}
\end{table*}

Representative tasks illustrate the scope of each domain.
In the office domain, one task asks the agent to re-phase a leak-loss pilot of a water utility and to decide on retention-fund approvals, delivering a workbook and a memo, and another asks it to rebuild a configuration-change flowchart from native presentation shapes and to reconcile review evidence across revisions.
In the communication domain, one task asks the agent to track the scope freeze, blocker decisions, and release candidates of an Apache Kafka release over nine rounds and 47 emails in five structured tables, and another asks it to track multi-state power-policy commitments and their fulfillment through a personal inbox of 200 emails.
In the research domain, one task revises an energy-indicator brief after reviewer feedback switches the main indicator, so that earlier conclusions must be qualified or retracted, and another migrates an earthquake-catalog database pipeline to a new major version of its database toolkit while preserving corrections and research annotations.
In the engineering domain, one task requires running a 28-day building energy baseline, fitting a thermal model from its results, designing a feeder-constrained heating controller and procurement decision, and redoing them after an authority bulletin cuts the budget, and another requires parametric CAD models and machining toolpaths for bridge members together with masking a moving weld seam across 900 video frames, before and after a supplier event.
In the business domain, one task continues a monthly net-settlement bridge for sales cancellations in a spreadsheet and an ERP system over three stages, and another revises a public spending dashboard over nine stages that add chart corrections, concentration and sensitivity analyses, and a year-over-year comparison.
In the media domain, one task adapts an open 3D animation scene into a short opening film and delivers both the rendered video and the native scene file, and another produces an onboarding presentation and a workstation sheet for a wire-harness workshop from public standards.

\subsection{Challenge Annotation}
\label{sec:app-challenges}

Each task is annotated with the six challenges of Table~\ref{tab:challenges} from its complete package, including the instruction, the instructions of every stage, the initial workspace, and the contract.
For every challenge, the annotator applies the criterion of Section~\ref{sec:challenges} and records the concrete feature of the task that makes the challenge necessary, such as an inherited workbook that must be preserved, a notice that arrives after the first submission, or a deliverable that must stay consistent with records in another application.
Annotations are reviewed for consistency across domains, and a challenge is assigned only when its evidence is concrete.

\begin{table*}[htbp]
\centering
\caption{The six cowork challenges targeted by FrontierCowork and the number of tasks that exhibit each challenge. A task exhibits a challenge only if the easier variant, which removes the challenge, would make the task clearly easier. Tasks may exhibit several challenges.}
\label{tab:challenges}
\small
\setlength{\tabcolsep}{3.6pt}
\renewcommand{\arraystretch}{1.18}
\begin{tabularx}{\textwidth}{>{\raggedright\arraybackslash}p{2.55cm}>{\raggedright\arraybackslash}X>{\raggedright\arraybackslash}p{2.45cm}rrrrrrr}
\toprule
Challenge & What the agent must do & Easier variant & Off. & Com. & Res. & Eng. & Bus. & Med. & All \\
\midrule
\multicolumn{10}{l}{\textit{Temporal}} \\[1pt]
Work continuation & Build on or repair an existing work product without breaking what must be kept & Starting from scratch & 9 & 0 & 8 & 12 & 18 & 19 & \textbf{66} \\
Long-horizon execution & Keep intermediate results valid over long runs in which earlier results drive later steps & Re-fetching the original inputs at any time & 17 & 26 & 15 & 16 & 20 & 20 & \textbf{114} \\
Dynamic adaptation & Revise finished work when requirements or the environment change after work has started & Stating all requirements up front & 0 & 26 & 6 & 12 & 11 & 0 & \textbf{55} \\
\midrule
\multicolumn{10}{l}{\textit{Spatial}} \\[1pt]
Requirement discovery & Infer the operative goal, rules, or scope from vague, conflicting, or stale materials & A fully specified instruction & 17 & 26 & 8 & 6 & 11 & 1 & \textbf{69} \\
Cross-resource dependency & Keep information consistent across distinct files, applications, and deliverables & Merging everything into one file & 19 & 26 & 15 & 16 & 20 & 20 & \textbf{116} \\
Multimodal grounding & Read images, renders, or frames and check visual deliverables by inspection & Replacing images with text descriptions & 9 & 0 & 8 & 4 & 4 & 20 & \textbf{45} \\
\midrule
\multicolumn{3}{l}{Tasks} & 22 & 26 & 16 & 16 & 20 & 20 & 120 \\
\multicolumn{3}{l}{Challenges per task} & 3.2 & 4.0 & 3.8 & 4.1 & 4.2 & 4.0 & 3.9 \\
\bottomrule
\end{tabularx}
\end{table*}

\section{Execution and Scoring Details}
\label{sec:app-execution}

Table~\ref{tab:domains} summarizes the deliverables, workflow structure, and scoring approach for each domain.

\begin{table*}[!htbp]
\centering
\caption{The six FrontierCowork domains: task counts, deliverables, workflow structure, and scoring. Research tasks with multiple stages use the final-stage score; communication and business tasks average scores across rounds or stages.}
\label{tab:domains}
\small
\setlength{\tabcolsep}{5pt}
\renewcommand{\arraystretch}{1.18}
\begin{tabularx}{\textwidth}{@{}l r >{\raggedright\arraybackslash}X >{\raggedright\arraybackslash}p{3.05cm} >{\raggedright\arraybackslash}p{3.25cm}@{}}
\toprule
\textbf{Domain} & \textbf{Tasks} & \textbf{Deliverables} & \textbf{Workflow} & \textbf{Scoring} \\
\midrule
Office & 22 & Spreadsheets, documents, slides & Single stage & Programmatic; pass/fail \\[3pt]
Communication & 26 & Structured records, handoff briefs & 5--9 rounds & Record matching; continuous \\[3pt]
Research & 16 & Tables, reports, figures, pipelines & 1--2 stages & Programmatic; mostly pass/fail \\[3pt]
Engineering & 16 & Repaired artifacts, release decisions & Single stage; operator events & Programmatic; pass/fail \\[3pt]
Business & 20 & Analysis workbooks, ERP records & 3--16 stages & Programmatic; stage pass/fail, averaged \\[3pt]
Media & 20 & Editable projects, media exports & Single stage & Rubric review; continuous \\
\midrule
\textbf{Total} & \textbf{120} & & & \\
\bottomrule
\end{tabularx}
\end{table*}

\subsection{Evaluated Systems and Protocol}
\label{sec:app-systems}

Table~\ref{tab:main} gives the full system names and their harnesses. All systems use the highest reasoning effort they support: Astra runs in Codex and the other nine models run in Claude Code. Each system is evaluated once on each of the 120 tasks. Infrastructure failures are rerun, while failures attributable to the agent remain in the evaluation. Media submissions are reviewed by GPT-6.1 Sol at high reasoning effort with the evaluated system's identity hidden.
The main comparisons therefore concern complete model--harness systems under the reported configuration; they do not isolate the effect of the model or estimate variation across repeated runs.

\subsection{Task Package Format}
\label{sec:app-package}

Every task is distributed as a directory in the Harbor task format~\cite{harbor2026}.
The directory contains an agent-facing instruction, a configuration file, an environment specification with the container definition and any task-specific services, a reference solution, and a verifier that writes the task reward.
The configuration file declares the agent time budget, the network policy, the user under which the agent runs, the container image and its resource limits, the verifier time limit, and the artifacts that are collected after the run.
The verifier always runs in a separate container from the agent, and the reference solution, the verifier, and all reference values are absent from the agent-facing workspace.
Multi-stage tasks additionally contain one subdirectory per stage with its own instruction, delivered materials, and stage verifier, and the materials of a stage are placed in the workspace only when that stage begins.
Communication tasks include an email workbench that serves the archive round by round through a local tool server, engineering tasks include an operator service that issues notices and receives the final release or hold decision, and business tasks include a spreadsheet application or an ERP system whose state persists across stages.
Media tasks contain the commission, the task-specific rubric, and an assessment contract in place of a programmatic verifier.
Container resources range from one CPU and 1\,GB of memory for communication tasks to four CPUs and 16\,GB of memory for media tasks, which must render video and 3D scenes.

\subsection{Task-Specific Scoring Criteria}
\label{sec:app-scoring}

\paragraph{Office and engineering.}
Office and engineering verifiers are binary, and a task scores 1 only if every check of the contract passes.
Office verifiers open the delivered files in their native applications, recompute formulas after editing inputs in place, and check values, structure, formatting, and protected regions.
Engineering verifiers replay the delivered artifacts in the native toolchain, such as a layout flow, a simulator, a CAD kernel, or a build system, and the score is set to zero if the workspace is modified after the final replay.

\paragraph{Research.}
The research domain contains sixteen tasks, most of which receive a pass or fail score based on all checks of the contract.
The other tasks award the fraction of passed checks; where specified, checks whose outcome cannot be decided from the delivered evidence are excluded from the denominator.
For multi-stage research tasks, the score is taken from the final stage, whose deliverable must incorporate all earlier stages.

\paragraph{Communication.}
In every round, the agent maintains structured tables and saves a handoff snapshot with a brief, and a round without a saved snapshot or with a brief shorter than 200 characters scores zero.
Submitted rows are matched to reference rows by their canonicalized key fields, where canonicalization folds case, whitespace, and declared aliases.
A matched row earns full credit if every scored field agrees with the reference or an accepted alternative, half credit if all critical fields agree but a general field does not, and no credit otherwise, and missing or duplicated rows earn no credit.
Fields are compared by type, so that enumerations ignore case and stated synonyms, numbers are compared numerically, timestamps are compared as instants in UTC, names and text are compared after alias normalization, and commit hashes match by prefix.
Unknown values are represented as blank, and a blank field matches only a blank reference, which rewards leaving a field empty over guessing it.
Every submitted row that matches no reference row adds one to the denominator, so the round score is the credit earned divided by the number of reference rows plus the number of unmatched submitted rows, and the task score is the mean over rounds.

\paragraph{Business.}
Each stage of a business task has its own verifier that checks the analysis workbook, the handoff artifacts, or the records in the ERP system, and the stage is passed only if all of its checks pass.
The task score is the mean of the stage scores, and a failed stage does not end the task, so the agent continues to later stages and can still earn credit for them.

\paragraph{Media.}
\label{sec:app-media}
Each media task has its own rubric with five dimensions scored from 0 to 3, for a maximum of 15 points.
Exactly one dimension concerns factual fidelity to the commission and its sources, and the other dimensions typically cover visual expression and finish, continuation of the native project for future editing, and task-specific qualities such as audience communication or narrative.
Within each dimension, a score of 1 denotes a usable result, 2 denotes substantive professional quality, and 3 denotes concrete value beyond a strong result.
The task score is the total divided by 15.
The reviewer is GPT-6.1 Sol with high reasoning effort, run in a fresh context for every submission.
It receives the commission, the rubric, general guidance on artistic quality, the assessment contract, the materials visible to the agent, and a read-only copy of the submission, and it receives neither the identity of the evaluated system nor any other score.
The reviewer inspects the exported products directly, renders its own evidence from copies, such as timestamped frame sequences for video, and tests the native files by opening, editing, saving, and reopening copies of them.
A deliverable that is missing, empty, or cannot be opened scores zero.
If the reviewer cannot complete a review, the submission is reviewed once more, and a submission that still cannot be reviewed is treated in the same way as a run without a valid score.

\section{Supplementary Experimental Results}
\label{sec:app-results}

\subsection{Complete Challenge Scores}
\label{sec:app-profile}

Table~\ref{tab:challenge-profile} reports the complete numerical results underlying the challenge profiles in Figure~\ref{fig:challenge-radar}.

\begin{table*}[!htbp]
\centering
\caption{Mean score (0 to 100) of each system on the tasks that exhibit each challenge. Tasks is the number of tasks that exhibit the challenge; a task may exhibit several challenges, so the subsets overlap and differ in their domain composition. Bold values denote the highest value in each row.}
\label{tab:challenge-profile}
\small
\setlength{\tabcolsep}{4pt}
\begin{tabular}{lrrrrrrrrrrr}
\toprule
\textbf{Challenge} & \textbf{Tasks} & \textbf{Astra} & \textbf{Opus} & \textbf{Fable} & \textbf{Sonnet} & \textbf{DeepSeek} & \textbf{Hy4} & \textbf{Kimi} & \textbf{GLM} & \textbf{Doubao} & \textbf{Qwen} \\
\midrule
Work continuation & 66 & \textbf{52.5} & 46.5 & 49.6 & 42.5 & 43.4 & 37.4 & 31.1 & 29.1 & 25.2 & 17.4 \\
Long-horizon execution & 114 & \textbf{58.2} & 48.5 & 42.0 & 37.5 & 37.0 & 27.8 & 28.5 & 24.5 & 24.4 & 18.0 \\
Dynamic adaptation & 55 & \textbf{56.0} & 52.4 & 41.1 & 42.7 & 37.0 & 25.1 & 34.2 & 25.3 & 27.8 & 19.1 \\
Requirement discovery & 69 & \textbf{55.1} & 40.8 & 38.0 & 38.4 & 27.3 & 22.2 & 26.1 & 25.7 & 18.8 & 10.8 \\
Cross-resource dependency & 116 & \textbf{58.0} & 47.6 & 42.1 & 38.6 & 37.2 & 29.0 & 28.0 & 25.8 & 24.9 & 16.9 \\
Multimodal grounding & 45 & \textbf{57.0} & 56.3 & 51.4 & 39.0 & 41.7 & 28.6 & 26.4 & 32.9 & 26.1 & 21.8 \\
\bottomrule
\end{tabular}
\end{table*}

\subsection{Detailed Domain Analyses}
\label{sec:app-domain-results}

\paragraph{Communication task forms.}
Within communication, tasks that require deep coordination of a single transaction across a long email archive are the hardest, with an average of 21.3 across systems, compared with 29.9 for release coordination on open-source mailing lists and 27.1 for inboxes that interleave several projects.
The gap is consistent across systems, as Astra scores 41 on single-transaction coordination against 48 on release coordination.
Single-transaction archives require the agent to decompose each message into many fine-grained records and to keep earlier records consistent as later messages revise them, which combines long-horizon state tracking with continual adaptation to revised facts.

\paragraph{Media completion and task categories.}
In media design, every task requires the agent to inspect source images, renders, or frames and to deliver an edited native project together with its exported media, and a deliverable that is missing, empty, or cannot be opened, including one that is not finished within the budget, receives zero.
Incomplete deliveries vary widely across systems, occurring in at most two of the twenty media tasks for Fable, DeepSeek, GLM, and Opus, but in four to seven tasks for Sonnet, Astra, Doubao, Kimi, Hy4, and Qwen, which often exhaust the budget while rendering.
Motion graphics, which constitute the largest media category and require rendering and exporting a finished video from a native project, show the same separation: Astra and Fable reach 47 to 48 and Opus and DeepSeek 40 to 41, whereas Doubao, Hy4, Qwen, and Kimi remain between 11 and 22.
Producing a complete, openable visual artifact thus remains a non-trivial capability that verifiers based on intermediate outputs would fail to capture.

\section{Resource Usage and Cost Accounting}
\label{sec:app-cost}
\subsection{Token Accounting and Cost Estimation}

Providers report prompt-side usage differently, since some count cached input within the input tokens and others report it separately.
For each model call we therefore split prompt-side usage into uncached input $n_{\mathrm{in}}$, cache reads $n_{\mathrm{cr}}$, and cache writes $n_{\mathrm{cw}}$, so that every prompt token is counted exactly once, and we record output tokens $n_{\mathrm{out}}$, including reasoning tokens.
The tokens reported in Figure~\ref{fig:resources} are $n_{\mathrm{in}} + n_{\mathrm{cr}} + n_{\mathrm{out}}$ averaged over tasks, since cache writes are not reported uniformly across providers.
The estimated cost of a call is
\begin{equation}
    c = \frac{p_{\mathrm{in}} n_{\mathrm{in}} + p_{\mathrm{cr}} n_{\mathrm{cr}} + p_{\mathrm{cw}} n_{\mathrm{cw}} + p_{\mathrm{out}} n_{\mathrm{out}}}{10^{6}},
\label{eq:cost}
\end{equation}
where $p_{\mathrm{in}}$, $p_{\mathrm{cr}}$, $p_{\mathrm{cw}}$, and $p_{\mathrm{out}}$ are the official list prices in US dollars per million tokens.
Because providers do not report cache writes uniformly, the estimates in Table~\ref{tab:prices} set $n_{\mathrm{cw}} = 0$ and are therefore lower bounds for providers that charge for cache writes.
Table~\ref{tab:prices} reports the list prices, the mean token usage per task, and the resulting estimated cost of every system.
Costs computed in this way are estimates at list prices and may differ from the amounts charged through a particular gateway.

\begin{table*}[!ht]
\centering
\caption{Token usage and estimated inference cost of each system on FrontierCowork, sorted by cost. Prices are official API list prices on the standard tier collected from the vendor pricing pages on 2026-11-25, with prices listed in Chinese yuan converted at 6.7111 yuan per dollar. Token counts are averaged over the runs behind the reported scores, with cached input separated from uncached input and output including reasoning tokens; cache writes are not reported uniformly by providers and are excluded.}
\label{tab:prices}
\small
\setlength{\tabcolsep}{4pt}
\resizebox{\textwidth}{!}{%
\begin{tabular}{lrrrrrrrr}
\toprule
 & \multicolumn{3}{c}{\textbf{List price (\$ per M tokens)}} & \multicolumn{3}{c}{\textbf{Mean tokens per task}} & \multicolumn{2}{c}{\textbf{Estimated cost (\$)}} \\
\cmidrule(lr){2-4}\cmidrule(lr){5-7}\cmidrule(lr){8-9}
\textbf{Model} & \textbf{Input} & \textbf{Cache read} & \textbf{Output} & \textbf{Input (M)} & \textbf{Cache read (M)} & \textbf{Output (K)} & \textbf{Per task} & \textbf{120 tasks} \\
\midrule
Claude Fable 5.1 & 10.00 & 0.250 & 50.00 & 0.28 & 7.07 & 81 & 8.60 & 1{,}032 \\
GPT-6.2 Astra & 8.00 & 0.400 & 40.00 & 0.58 & 3.73 & 61 & 8.58 & 1{,}029 \\
Kimi K3.1 Pro & 2.00 & 0.200 & 10.00 & 0.66 & 17.73 & 71 & 5.58 & 670 \\
Claude Opus 5.5 & 4.00 & 0.200 & 20.00 & 0.27 & 4.43 & 91 & 3.78 & 453 \\
Qwen3.8 Max & 1.79 & 0.224 & 5.36 & 0.48 & 3.81 & 136 & 2.44 & 293 \\
Claude Sonnet 5.5 & 2.00 & 0.100 & 10.00 & 0.28 & 4.11 & 116 & 2.13 & 255 \\
Doubao Seed Evolving & 0.894 & 0.179 & 4.47 & 0.33 & 7.04 & 112 & 2.05 & 246 \\
Tencent Hy4 & 0.894 & 0.045 & 2.68 & 0.36 & 9.26 & 153 & 1.15 & 138 \\
GLM-5.3 Flash & 0.150 & 0.030 & 0.500 & 0.30 & 14.17 & 139 & 0.54 & 65 \\
DeepSeek V4.1 Flash & 0.300 & 0.006 & 1.20 & 0.29 & 15.55 & 135 & 0.34 & 41 \\
\bottomrule
\end{tabular}%
}
\end{table*}

\section{Additional Case Studies}
\label{sec:app-cases}

We complement the two cases in Section~\ref{sec:failure-analysis} with two further tasks involving engineering revisions and a visual data audit.
Across the four cases, the systems that succeed and the systems that fail can be separated by a concrete artifact in their delivered work or their trajectories.
Each task carries four of the six cowork challenges, and together the four tasks cover all six, so the cases show how the challenges compound within one assignment rather than isolating any single one.
Each case gives its identifier in the released package and lists the challenges it carries.
Every claim is traced to a verifier output, a delivered file, or a logged agent step, and the figures reproduce these artifacts directly.

\begin{figure*}[!ht]
\centering
\includegraphics[width=\textwidth]{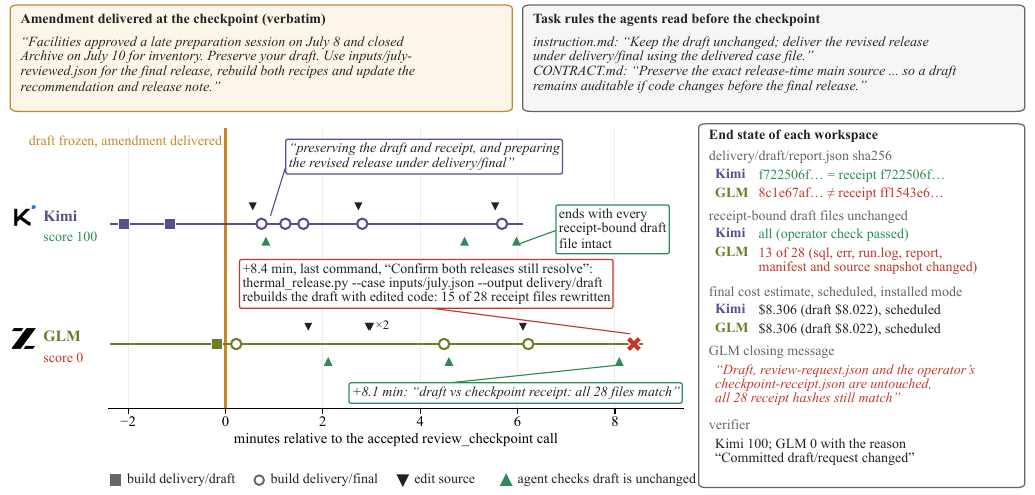}
\caption{Museum Cooling Migration (engineering 33). Kimi and GLM around the checkpoint, reconstructed from the agent logs with time in minutes relative to each agent's accepted checkpoint call.
Markers show builds into the draft and final release directories, edits to the planner source after the checkpoint, and the agents' own checks that the draft is unchanged.
Quotes from the amendment, the task files and the agents are verbatim.
The right panel gives the end state read from the delivered workspaces and the verifier.}
\label{fig:case-adapt}
\end{figure*}

\subsection{Museum Cooling Migration}
\label{sec:case-adapt}
\noindent\textit{Challenges: work continuation, long-horizon execution, dynamic adaptation, and cross-resource dependency.}

Engineering task 33 asks the agent to repair a Python cooling planner, run four EnergyPlus simulations, and release a draft before a review checkpoint.
The checkpoint freezes the draft with a receipt of file hashes and delivers an amendment that says \textit{Preserve your draft} and asks for a final release from a revised calendar, and the instruction already reads \textit{Keep the draft unchanged; deliver the revised release under delivery/final}.
Applying the amendment was the easy part, since Astra, Opus, Fable, Kimi and GLM all report the same final cost of \$8.306.

The difficulty is that work continues after the checkpoint, and once the planner source changes, rebuilding the draft can no longer reproduce the frozen bytes.
Kimi edited the source three more times but rebuilt only the final release (Figure~\ref{fig:case-adapt}).
GLM edited it four times and confirmed at 8.1 minutes that \textit{all 28 files match}, then eighteen seconds later ran a last check, \textit{Confirm both releases still resolve}, that rebuilt the draft with the edited code and rewrote 15 of the 28 receipt-bound files.
It closed by reporting that \textit{all 28 receipt hashes still match}, and the verifier scored it 0 with \textit{Committed draft/request changed}, while Kimi scored 100.

Carrying a change into a new revision includes protecting the old one, and a habit of rerunning everything to confirm success can overwrite the very state the change asked the agent to keep without the agent noticing.

\begin{figure*}[!ht]
\centering
\includegraphics[width=\textwidth]{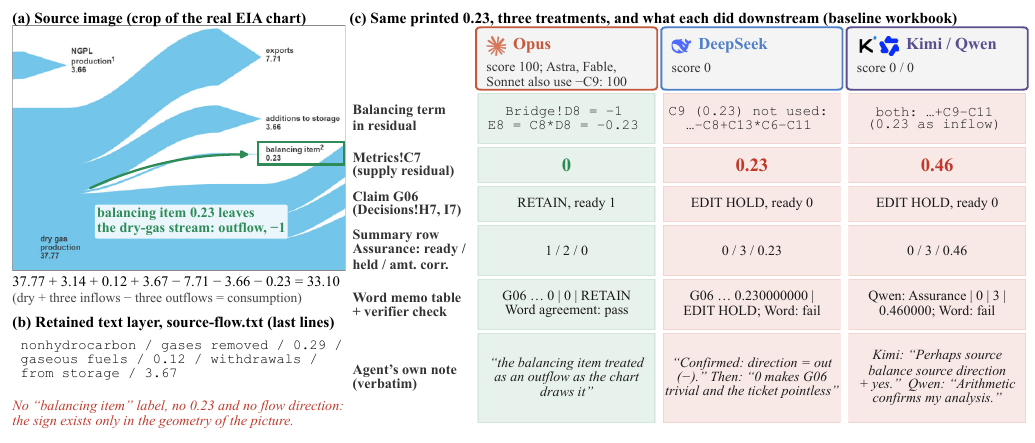}
\caption{Natural Gas Restatement Audit (CB4-MM-012). One printed value and three treatments of its sign. (a) Crop of the real EIA 2024 natural gas flow diagram given to the agents; the balancing item branch (boxed) leaves the dry-gas stream, which makes it an outflow and closes the bridge to published consumption of 33.10. (b) The last lines of the retained text layer, which carry no balancing item label, value, or direction. (c) Cells from the delivered workbooks after LibreOffice recalculation (formulas shown without the \texttt{Source!} prefix), the corresponding row of the Word handover with the verifier's numeric agreement result, and verbatim notes from the agents' trajectories. Green marks values that match the reference and red marks values that do not.}
\label{fig:case-diagram}
\end{figure*}

\subsection{Natural Gas Restatement Audit}
\label{sec:case-diagram}
\noindent\textit{Challenges: long-horizon execution, requirement discovery, cross-resource dependency, and multimodal grounding.}

Office task CB4-MM-012 asks for a live review workbook and a Word handover that audit twelve draft claims against the real EIA natural gas flow diagram for 2024, with a supply bridge that includes the balancing item \textit{with source directions}.
The value 0.23 is legible, but its sign lives only in the geometry of the chart, since the text layer carries neither the label, the value, nor a direction (Figure~\ref{fig:case-diagram}a and b).
As an outflow it closes the bridge exactly at 33.10, so the residual claim G06 of 0 must be retained.

All eight systems that delivered a workbook transcribed 0.23, and Astra, Opus, Fable and Sonnet subtracted it and scored 100.
DeepSeek wrote \textit{Confirmed: direction = out ($-$)} and called a zero residual \textit{the faithful reading}, then dropped the term because \textit{0 makes G06 trivial and the ticket pointless}.
Kimi added it as an inflow, and Qwen, whose plan subtracted it because \textit{in chart it's an outflow}, wrote \texttt{+BALANCE} into its check script and concluded \textit{Arithmetic confirms my analysis}.
The wrong residual flipped G06 to EDIT HOLD and carried into the summary tables and the Word memo (Figure~\ref{fig:case-diagram}c), and all three scored 0, Kimi and Qwen with an additional column error in Decisions.

The decisive skill was not perceiving the branch but trusting it: two failing systems held the correct sign and replaced it, one with a guess about the author's intent and one with an unchecked line of code.

\end{document}